%% file: neuraldep.tex
\documentclass{article}

\usepackage[utf8]{inputenc} % allow utf-8 input
\usepackage[T1]{fontenc}    % use 8-bit T1 fonts

\usepackage{times}
\usepackage{graphicx} % more modern
\usepackage{subfigure} 

\usepackage{natbib}

\usepackage{algorithm}
\usepackage{algorithmic}

\usepackage{hyperref}

\usepackage{booktabs}       % professional-quality tables
\usepackage{amsfonts}       % blackboard math symbols
\usepackage{amsmath}
\usepackage{amssymb}
\usepackage{nicefrac}       % compact symbols for 1/2, etc.
\usepackage{microtype}      % microtypography
\usepackage{graphicx}

\usepackage[accepted]{icml2017}

\icmltitlerunning{Fully-neural Dependency Parsing}

\begin{document} 

\twocolumn[
\icmltitle{Read, Tag, and Parse All at Once,\\ or Fully-neural Dependency Parsing}



% you can specify symbols, otherwise they are numbered in order
\icmlsetsymbol{equal}{*}

\begin{icmlauthorlist}
\icmlauthor{Jan Chorowski}{UW}
\icmlauthor{Michał Zapotoczny}{UW}
\icmlauthor{Paweł Rychlikowski}{UW}
\end{icmlauthorlist}

\icmlaffiliation{UW}{Institute of Computer Science, University of Wrocław, Poland}

\icmlcorrespondingauthor{Jan Chorowski}{jan.chorowski@cs.uni.wroc.pl}

\icmlkeywords{machine learning, dependency parsing}

\vskip 0.3in
]



\printAffiliationsAndNotice{\icmlEqualContribution} % otherwise use the standard text.
%\footnotetext{hi}

\begin{abstract} 
  We present a dependency parser implemented as a single deep neural
  network that reads orthographic representations of words and
  directly generates dependencies and their labels. Unlike typical
  approaches to parsing, the model doesn't require part-of-speech
  (POS) tagging of the sentences. With proper regularization and
  additional supervision achieved with multitask learning we reach
  state-of-the-art performance on Slavic languages from the Universal
  Dependencies treebank: with no linguistic features other than
  characters, our parser is as accurate as a transition-based system
  trained on perfect (manually provided) POS tags.
\end{abstract} 

\section{Introduction}

The ability to communicate using natural language is one of the
long-term goals of artificial intelligence. Moreover, due to the huge
amount of natural language texts there is a growing need to
develop effective algorithms to handle them in a satisfactory
manner. In the last decades one could observe a shift of focus from
linguistics to statistical text analysis and more recently to machine
learning systems and neural networks. 

Deep learning methods have led to many breakthrough in NLP tasks, such as language
modeling \citep{mikolov_recurrent_2010}, machine translation
\citep{bahdanau_neural_2014,sutskever_sequence_2014}, caption
generation \citep{xu_show_2015}, question answering
\citep{sukhbaatar_end--end_2015}, speech regognition, POS-taggers and
so on.  

Finding the syntactical structure of sentences is one of the
essential needs in natural language text analysis. Parsing is a key
component required for automated natural language
understanding. Virtually all NLP task could benefit from having a good
quality parse tree for analyzed sentence.  

In this contribution we build a deep learning dependency parser that
operates directly on characters. The parser brings together many
recent ideas. On the one hand, following
\cite{kiperwasser_simple_2016,zhang_dependency_2016,dozat_deep_2016}
we replace the traditional stack-based parser architecture with a deep
recurrent neural encoder followed by a scoring network tasked with
selecting the root words. However, similarly to
\cite{kim_character-aware_2015,ballesteros_improved_2015} we also
remove the dependency on hand-crafted word annotations, such as
part-of-speech (POS) tags, using instead only raw characters. Despite
this lack of feature engineering, our parser achieves good performance
on morphologically rich Slavic languages. Furthermore, we show how
proper regularization and multi-task learning can greatly reduce
overfitting and make our model competitive even on languages with
limited training resources.

Our implementation is available at
\url{http://github.com/mzapotoczny/dependency-parser}.

\section{Description of the Model}

\begin{figure*}[t]
  \centering
  \includegraphics[width=\linewidth]{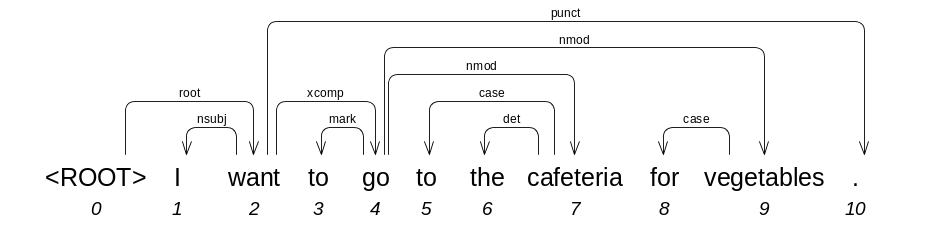}
  \caption{Dependency tree for sample sentence from English UD treebank.} 
  \label{fig:ilovenips}
\end{figure*}

\begin{figure}[!t]
  \centering
  \begin{minipage}{0.5\textwidth}
    \resizebox{!}{2.0\textwidth}{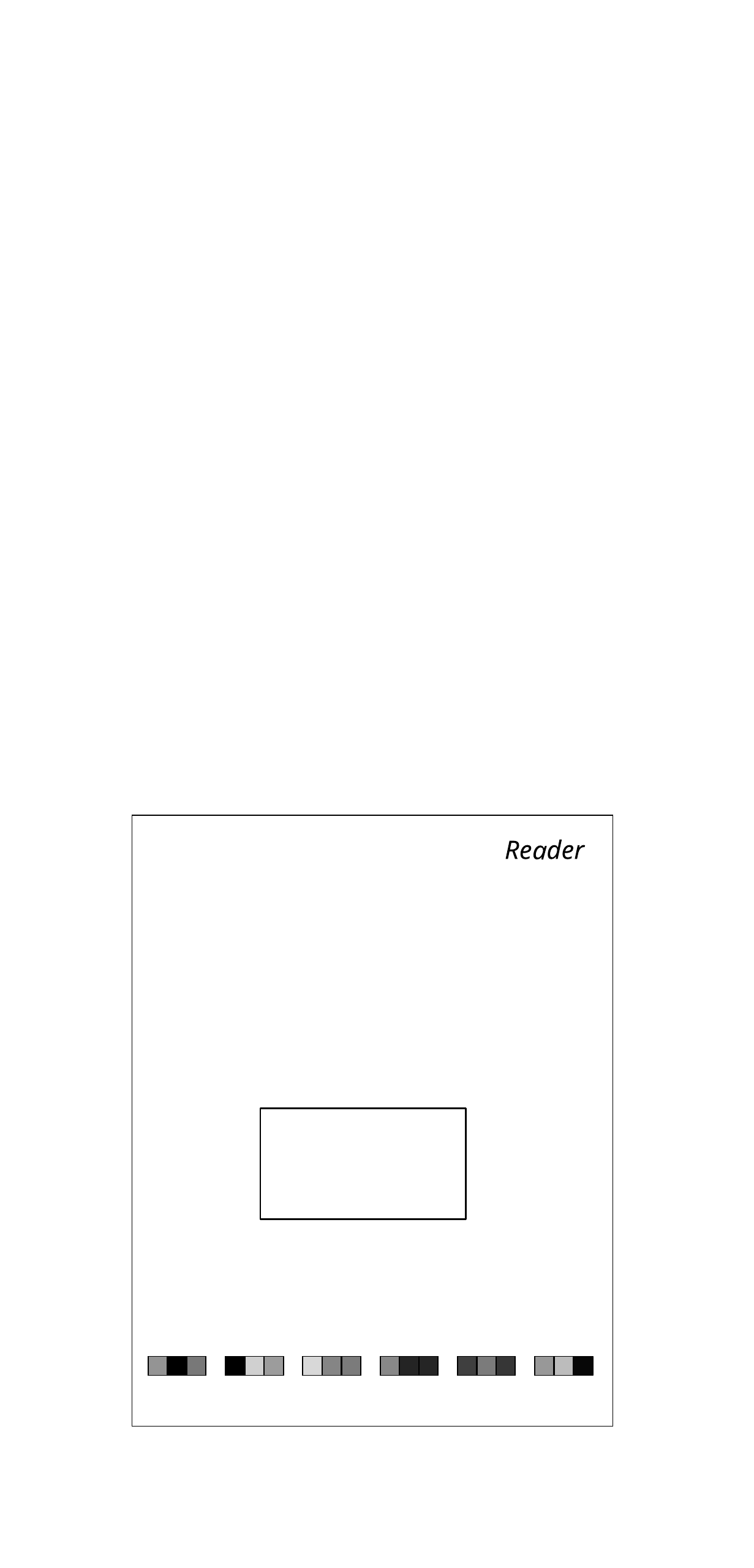}
  \end{minipage}
  \caption{The proposed model architecture.} 
  \label{fig:architecture}
\end{figure}

A dependency parser reads a sentence and finds a set of dependencies,
that are triples composed of a \emph{head} word, a \emph{dependent}
word, and a \emph{label} describing the dependency type. Each word has
exactly one head, with one word in the sentence (typically the
verb) having an artificial <ROOT> token attached as its head. Therefore the set of
dependencies can be interpreted as an oriented tree linking words of
the sentence. Please see Figure~\ref{fig:ilovenips} for an exemplary
dependency tree.

Our dependency parser is implemented as a single neural network with three
parts, as depicted in Figure~\ref{fig:architecture}. First, the
\emph{reader} subnetwork finds word
embeddings based on their orthographic representations using
convolutional and highway layers
\citep{kim_character-aware_2015,srivastava_highway_2015}. Second, a
bidirectional recurrent \emph{tagger} subnetwork puts the individual
words into their contexts \citep{schuster_bidirectional_1997}. Finally, the
\emph{parser} subnetwork uses the soft-attention mechanism to point
each word to its head
\citep{vinyals_pointer_2015,bahdanau_neural_2014}. Once the head is
found, it is used to compute the dependency label.

\subsection{Reader subnetwork}

The reader subnetwork is tasked with finding embeddings of words given
their orthographic representations. For many languages the spelling of
a word is a strong indicator of its grammatical function. Following
\citet{kim_character-aware_2015}, we use a convolutional filterbank
optionally followed by a few layers of nonlinear transformations. Each word $w$ is
represented by a sequence of its characters, fenced with special
beginning-of-word and end-of-word markers. We find low-dimensional
embeddings of characters and concatenate them to form a matrix
$C^w$. 

Next, the matrix $C^w$ is reduced to a vector of filter responses $R^w
\in \mathbb{R}^{\text{nf}}$, where \emph{$\text{nf}$} denotes the number of
filters. Each filter response is computed as:

\begin{equation*}
  R^w_i = \max(C^w \circledast F^i),
\end{equation*}

where $F^i$ is the $i$-th filter and $\circledast$ denotes convolution
over the length of the word. Intuitively, the convolutions act like
pattern matches that react to specific parts of the word. Furthermore,
the filters can differentiate between prefixes, suffixes and infixes
by reacting to the beginning- and end-of-word markers that are added to
each word.

%TODO: redo experiments without highway layers for other languages!
Finally, the reader applies a nonlinear transformation to filter
responses $R^w$. First, a linear transformation is used to reduce the
dimensionality of the representation. Then a stack of highway layers
\citep{srivastava_highway_2015} is applied to obtain the final word
embeddings $E^w$. 

\subsection{Tagging subnetwork}

The tagging subnetwork works on sequences of representations of words $E^w$
produced by the reader. It uses bidirectional recurrent layers (BiRNN) to put
them into a broader context
\citep{schuster_bidirectional_1997}. Specifically, we use GRU
recurrent units \citep{cho_properties_2014} to scan the sequence
forward and backward. The hidden representations are combined using
addition, and passed to another layer of recurrent units. 

The tagger can be trained based solely on the gradient signal
flowing into it from the parsing subnetwork. However, it is also
possible to branch off the signal from one of the BiRNN layers and
use it to predict part of speech (POS) tags of individual words. This
additional supervision typically helps to prevent overfitting of the
parser. In the experimental results section we present the impact of
explicit POS-tag training.

\subsection{Parsing subnetwork}\label{sec:parsing_net}

The parsing subnetwork has two objectives: first, to match dependent
words to their heads and second, to label each pair of matched words
with the proper dependency type.

We have chosen to use the pointer network \citep{vinyals_pointer_2015}
approach to find head words. For each sentence the parser obtains a
sequence $H_1, H_2, \ldots, H_n$ of vectors of word annotations
produced by the tagger. We prepend to this sequence a special vector
$H_0$ denoting the root word. This guarantees, that each word of the
original sentence has exactly one head word. To train the pointer
network we construct a probability distribution over possible head
word locations $l\in{0,1,\ldots,n}$.

First, for each word $w \in {1,2,\ldots n}$ we compute a score over
all possible locations $l$:

\begin{equation}
  s(w,l) = f(H_w, H_l),
\end{equation}

Where the \emph{scorer} $f(\cdot, \cdot)$ is implemented as a small
feed-forward neural network. 

The scores are normalized over locations using the SoftMax function to
$p(w,l)$, which are interpreted as the probabilities that the head of
word $w$ is at location $l$:

\begin{equation} \label{eq:loc_probs}
  p(w,l) = \frac{s(w,l)}{\sum_{l'=0}^n s(w,l')}
\end{equation}

Finally, the dependency \emph{label} is computed by a small Maxout network
\citep{goodfellow_maxout_2013} using the annotations of both the head
$H_h$ and the dependent $H_w$ words. We have analyzed two variants:
\begin{enumerate}
\item a \emph{soft attention} labeler, which uses the expected
  annotation of the head word computed under the probability
  distribution given by eq.~\eqref{eq:loc_probs} $H_h = \sum_lp(w,l)H_l$;
\item a \emph{hard attention} labeler, which uses the annotation $H_h$
  taken at the location of the correct head word. During training, the
  hard attention labeler uses the ground-truth head location while
  during evaluation we choose the most likely location according to
  \eqref{eq:loc_probs}. 
\end{enumerate}

We have briefly experimented with adding a recurrent hidden state to
the computation of scores $s$. The recurrent state was updated after processing each
word of the sequence. However, experiments have show little benefits
of this additional computation.

\subsection{Training criterion}

The network receives training signal from three sources:
\begin{enumerate}
\item The negative log-likelihood loss on predicting dependency labels
  $L_l$. With the soft attention labeler this loss is backpropagated
  through the entire network and theoretically could be used to train
  the entire network. With the hard attention labeler this error is
  not backprogated to the scorer.
\item The negative log-likelihood loss on finding proper head words $L_s$ by the scorer. This
  loss is backpropagated through the reader and tagger subnetworks,
  but not through the labeler.
\item The optional POS-tagging negative log-likelihood loss $L_t$. This loss
  is backpropagated only thorough a few layers of the tagger and
  through the reader.
\end{enumerate}

The final loss is computed as a linear combination of the individual
losses:

\begin{equation}
  L = \alpha_lL_l + \alpha_s L_s + \alpha_t L_t
\end{equation}

\subsection{Parsing algorithm}

At its core, the network produces, for each pair of words, scores that
reflect the probability that the words form a dependency. These scores
can be used to construct a parse tree by finding a set of dependencies that
satisfy some constraints (exactly one word is dependent on the root
token, there are no cycles, the tree is
projective). 

However, we have found that at the end of training the scores computed
by eq.~\eqref{eq:loc_probs} typically lead to a very peaked
probability distribution that is concentrated on just a single
location. Therefore good results are obtained with a greedy parsing
strategy that for each word simply chooses the best scoring
parent. Only approximately 0.5\% of the parses obtained by this
procedure have cycles, so using Chu-Liu-Edmonds
\citep{edmonds_optimim_1966} maximum spanning arborescence algorithm (which
deletes cycles) gives only a subtle improvement and is not presented
in the Table \ref{tab:results}.

\section{Related work}

There are two basic views on syntactic structure of the sentence:
\begin{itemize}
  \item constituent based, where words are organized in nested constituents
  \item dependency based, where words are connected by dependency relation
\end{itemize}

This work focuses on the dependency parsing. We believe that currently
two approaches are the most important ones: transition and graph
based. A transition based parser aims to predict the best parser
action (such as moving the word to stack or add a dependency between
current word and the word on a stack) looking at some features
\citep{nivre_algorithms_2008}. A
graph based parser finds the structure which maximizes a global score
while preserving some constraints (i.e. forces the output to be well
formed tree). Recently, deep neural networks were used wih a great
success in dependency parsing, both transition
\citep{chen_fast_2014,dyer_transition-based_2015,kiperwasser_simple_2016,andor_globally_2016,ballesteros_improved_2015} and graph
\citep{pei_effective_2015} based. Our parser is most similar to the
graph-based variant of
\citep{kiperwasser_simple_2016}. However, similarly to
\citep{ballesteros_improved_2015} we replace the word embeddings and
POS tags
with our reader subnetwork thus reducing the need for feature
engineering, which is an important
aspect of parser construction which requires knowledge of
linguistics. Powerful learning techniques reduce the burden of this
somewhat language-specific work.

Our parsing network brings together ideas from many recent
contributions. \citet{ling_finding_2015} successfully applied
character-based word embeddings computed with small BiRNNs (another
possible implementation of our ``reading'' subnetwork) to POS-tagging
and language modeling with recurrent networks. A similar mechanism was
employed in \citep{ballesteros_improved_2015} in a character-based
shift-reduce dependency parser. The character-based
word embeddings that we have used were described by
\citet{kim_character-aware_2015}. They were extensively analyzed and
compared against ones computed with BiRNNs by
\citet{jozefowicz_exploring_2016} in a large language-modeling study. 

A purely neural constituency parser was shown by
\citet{socher_parsing_2011}. It built a parse tree by repeatedly joining words
or subtrees using a recursive network. Later, 
\citet{vinyals_grammar_2014} have shown that good constituency
parsers can be created by learning to ``translate'' between a given sentence and
the linearization of its parse tree. The parser accessed the
source sentence through word embeddings, which were initialized
with Word2Vec \citep{mikolov_efficient_2013} and adapted during parser training.
We build on their work by directly
using the attention matrix as pointers into the source sentence
locations that correspond head words. This change greatly simplifies
the parser: there is no recurrent generator and no need for an
approximate search during evaluation. 

% Another example of neural networks parser (in this case graph-based)
% was proposed by \cite{}. Their parser computes the score for every
% potential small subgraph, and this score is used by other algoritm to
% find the final parse. They also describe how to use phrase level
% information in order to improve parsing. However, they rely on a
% external parser and don't use any information at the character
% level. Their best results require using second order features (such as
% sibling node and its local context). 

Our parser is unique in the fact that it requires virtually no data
engineering and the employed training criterion mimics the definition
of dependency parsing: it is trained to simply point head-words. The
model also has new and intriguing properties. Notably it is
confident enough in its predictions to allow for greedy creation of
parse trees. 

% To the best of our knowledge, we have built the first parser that is
% trained in an end-to-end manner reading directly characters of words
% and producing the parse tree, rather than relying on a separately
% trained tagger. This approach is especially beneficial morphologically
% rich languages for which the orthographic representation is
% meaningful. 

\section{Experimental Setup}

We have evaluated our parser on three languages, English, Czech, and
Polish from the Universal Dependencies (UD) v. 1.2
dataset \citep{nivre_universal_2015}. We have chosen this dataset
because of its wide availability\footnote{Unfortunately we couldn't
  access the more typically used CoNLL '09 shared task data because the
  licensing webpage was not operational during the preparation of
  this manuscript.} and because in the future we want to
investigate the possibility of cross-lingual training. While the
English treebank used in UD is rather small and non-standard,
treebanks for other languages are often the typical and standard
ones. In particular, we evaluate on Polish for which the UD project
uses the only polish treebank ``Składnica''
\citep{swidzinski_towards_2010} and on Czech for which UD uses the
large and standard ``Prague Treebank''
\citep{bejcek_prague_2013}. Properties of the dataset are gathered in
Table \ref{tab:datasets}.

\begin{table}[tb]
  \centering
  \caption{Dataset sizes}
  \label{tab:datasets}
  \begin{tabular}{ccc}
    language & \#tokens & \#sentences \\ \hline
    Czech & 1503k & 87.9k\\
    English & 255k & 16.6k\\
    Polish & 83.5k &  8.2k
  \end{tabular}
\end{table}

\paragraph{Model selection} We have conducted a hyperparameter search
on the Polish treebank, which is the smallest one. We
have used the Spearmint system to choose network layer sizes and
regularization hyperparameters \citep{snoek_practical_2012}. Based on
the experiments on Polish we have chosen three network configurations
that we have used for Czech and English. On all languages we have
trained the networks on the provided training splits and performed
model selection and early stopping on the development split. 

The best overall network used 1050 filters in the reader subnetwork
($50\cdot k$ filters of length $k$ for $k=1,2,\ldots, 6$) whose
outputs were projected into $512$ dimensions and transformed by $3$
Highway layers. The tagging network consisted of 2 BiRNN layers each
composed of $548$ GRU units whose hidden states were aggregated using
addition. Head words were pointed to by a small MLP with 384 $\tanh$
units in the hidden layer and the labeler used one hidden layer of 256
Maxout units, each using 2 pieces. For evaluation on the gold POS tags
provided with the treebanks we have used a reader that embedded the
base forms and each of the tag attributes into $192$ dimensions.

We have added a projection layer between filters and highway units in
the reader to speed the computations: we have discovered that
generally increasing the number of filters is beneficial, however with
a large number of filters the highway layers became a bottleneck (they
are twice more expensive to evaluate than standard fully connected
layers). Likewise, we have used a large dropout fraction in the BiRNN
encoder and we have decided to add, rather than concatenate the hidden
activations to reduce the input into the next recurrent layers.

\paragraph{Training procedure} All non-recurrent weights were
initialized from a normal Gaussian distribution with standard
deviation set to $0.01$, while the recurrent weights were
orthogonalized. Initial states of recurrent layers were learned. We
have used the AdaDelta \citep{zeiler_adadelta:_2012} learning rule
with parameters $\epsilon=10^{-8}$ and $\rho=0.95$. We have routinely
used an adaptive gradient clipping mechanism
\citep{chorowski_end--end_2014}. All runs were early stopped based on
the Unlabeled Attachment Score (UAS).

The primary training criterion was a linear combination of negative
log-likelihoods of proper head word detection (taken with a weight of $\alpha_s=0.6$) and dependency label
prediction (taken with a weight of $\alpha_l=0.4$). In experiments in which POS
tags were used as auxiliary training targets we have split the
POS tags into individual attributes and added their negative
log-likelihood costs with a weight $\alpha_t=1$.

\paragraph{Regularization} Polish and English treebanks are rather
small and proper regularization was crucial to achieve optimum
performance. We have obtained best results with Dropout
\citep{srivastava_dropout:_2014}. We have applied 20\% Dropout just
after the Reader subnetwork, 70\% after every BiRNN layer in the
tagger subnetwork \citep{pham_dropout_2013} and 50\% in the
labeler. In contrast to \citet{vinyals_grammar_2014} we have not used data
augmentations techniques. 

\section{Results}

Our parser has reached competitive performance with transition-based
dependency parsers, as demonstrated in Table~\ref{tab:results}. For
all datasets we report: the percentage of correctly labeled
dependencies (LA), the percentage of correctly attached heads
(Unlabeled Attachment Score, UAS), and the percentage of both
correctly attached heads and labels (Labeled attachment Score, LAS)
measured on the test set for the model that achieved the highest
performance on the development set. The results were computed using
the MaltEval tool.

We compare the performance of parsers in two regimes: first, to obtain
baselines we consider operation when the ground-truth (gold) POS tags
are given during inference. Second, we report results when the gold
POS tags are not available during inference and they either have to be
predicted using a separate tagger, or as in the case of our network,
the parser can directly refer to the spelling of words to infer their
grammatical function. The neural parser is competitive in both cases.

\begin{table*}[t]
  \centering
  \caption{Model performance on selected languages}
  \label{tab:results}
  \begin{tabular}{c|ccc|ccc|ccc}
     & \multicolumn{3}{c|}{Czech} & \multicolumn{3}{c|}{English} &
    \multicolumn{3}{c}{Polish} \\
    & LA & UAS & LAS & LA & UAS & LAS & LA & UAS & LAS \\ 
    \multicolumn{10}{c}{Gold POS tags used during inference} \\  \hline
    MaltParser  & 
    91.6 & 86 & 83 & 92.0 & 87.0 & 84.0 & 92.0 & 89.1 & 85.8
    \\
    \citep{straka_parsing_2015}  & - & 87.7 & 84.7 & - & 88.2 & 84.8 & - &
    89.8 & 85.5
    \\
    \citep{tiedemann_cross-lingual_2015}  & - & - & 85.7 &
    - & - & \textbf{85.7} & - & - & - \\ 
    NN (this work) & 
    \textbf{93.8} & \textbf{91.7} & \textbf{88} & 
    \textbf{92.3} & \textbf{88.6} & 85.1 & 
    \textbf{93.9} & \textbf{93.4} & \textbf{89.3} \\
    \multicolumn{10}{c}{Predicted POS tags used during inference} \\ \hline
    \citep{tiedemann_cross-lingual_2015} & - & - & 81.4 &
    - & - & 82.7 & - & - & - \\
    \multicolumn{10}{c}{No POS tags used during inference} \\  \hline
    NN words & 
    82.4 & 82.4 & 72.1 &
    85   & 81.9 & 74.7 & 
    73.8 & 74.6 & 61.6  \\
    NN chars, soft att. & 
    92.1 & \textbf{90.1} & 85.7 & % results/inprogress/lab110-01/dependency_norec_smaller_cz/pretraining_best.zip
    90 & 86.5 & 82.1 & % results/inprogress/sonata2/dep_en/pretraining_best.zip
    88.7 & 89.1 & 82.5 \\ %results/inprogress/lab110-11/dep_pl_lessclip/pretraining_best.zip
    NN chars, tags, soft att. & 
    89.5 & 89.6 & 82.8 & % lab110-02 results/inprogress/lab110-02/dependency_norec_smaller_cz_tags/pretraining_best.zip
    89.2 & 86.2 & 81.3 & % results/inprogress/sonata1/dep_en_tags/pretraining_best.zip
    89.3 & 90.4 & 83.9 \\ % results/inprogress/lab110-03/dep_cpw_opt_no_r_tags/pretraining_best.zip
    NN chars, tags, hard att. & 
    \textbf{92.6} & \textbf{90.1} & \textbf{86.7} & % results/inprogress/lab110-01/dependency_norec_smaller_cz_tags_l0_hard_restart/pretraining_best.zip
    \textbf{90.4} & \textbf{87.6} & \textbf{83.6} & % sonata2 local_storage/dependency_norec_spbest_en_tags_l0_hard/pretraining_best.zip
    \textbf{90.9} & \textbf{91.3} & \textbf{86} \\ % results/inprogress/lab110-17/dependency_norec_smaller_pl_tags_l0_hard/pretraining_best.zip

    \multicolumn{10}{p{0.9\textwidth}}{  Note: MaltParser results on
      Czech are sub-optimal because due to lack of computational
      resources we had to use a small dataset for parser optimization.}
  \end{tabular}
\end{table*}

\subsection{Baseline Models}

We have used MaltParser v. 1.8.1 tuned with MaltOptimizer
\citep{nivre_maltparser:_2005,ballesteros_maltoptimizer:_2012} on all 
information available in UD treebanks (gold POS). This gave us an
optimistic baseline, since during normal use POS tags will contain
errors due to the tagger. This error has been analyzed on UD v. 1.0 by
\citet{tiedemann_cross-lingual_2015}. As an additional optimized
baseline we include also results from \citet{straka_parsing_2015} that were
reported on the same version of UD treebanks that we use.

\subsection{Neural Parser with Golden Tags During Inference}

To compare our parser with the optimistic baseline, we have trained it
on gold POS tags. We have observed that best results were
obtained when the POS attributes were split and given to the network as
several categorical inputs. On Czech and Polish the neural network
improves the optimistic baseline error rates, while on
English the results are comparable.

\subsection{Neural Parser without POS Tags During Inference}

In the next experiment we have evaluated the network without POS tag
information during inference. When trained on individual words treated as discrete
entities, the performance of the parsing network has dropped
significantly, which can be seen in Table~\ref{tab:results}. One
solution, outlined by \citet{vinyals_grammar_2014} involves pre-training
word embeddings on a large corpus and using them in the input look-up
tables. However, we wanted to use the information present in the
spelling of each word and decided to use the character-based embedder
by \citet{kim_character-aware_2015}. Intuitively, in  morphologically
rich languages such as Czech or Polish the spelling of a
word conveys many hints about its grammatical function.

We have tested four variants of the networks: with and without an
auxiliary training objective consisting of predicting POS tags and
with the hard and soft attention in labelers
(c.f. Section~\ref{sec:parsing_net}). We have established that on
Polish for which has the smallest treebank multitask learning
increased the UAS score when the POS tag prediction used hidden states
of the penultimate BiRNN layer. On the larger datasets available on
Czech and English the extra supervision added by predicting POS tags
slightly decreases the results. However, on all languages the best
setup involved multitask learning and soft attention.

Using hard attention is also beneficial. We interpret
this fact as follows: with hard attention, the labeler always sees the
annotations of the correct head and dependent words, while with soft
attention the head annotation may refer to possibly many incorrect
words chosen by the attention mechanism. With hard attention gradients
from the labeler are backpropagated to the correct head word only,
which helps training. On the other hand, with soft attention the
gradients from the head nodes sometimes are backpropagated to
incorrect locations. This adds noise during training, but possibly
makes the labeler more robust to errors in localization of head
words.

\begin{figure}[tb]
  \centering
  \includegraphics[width=0.5\textwidth]{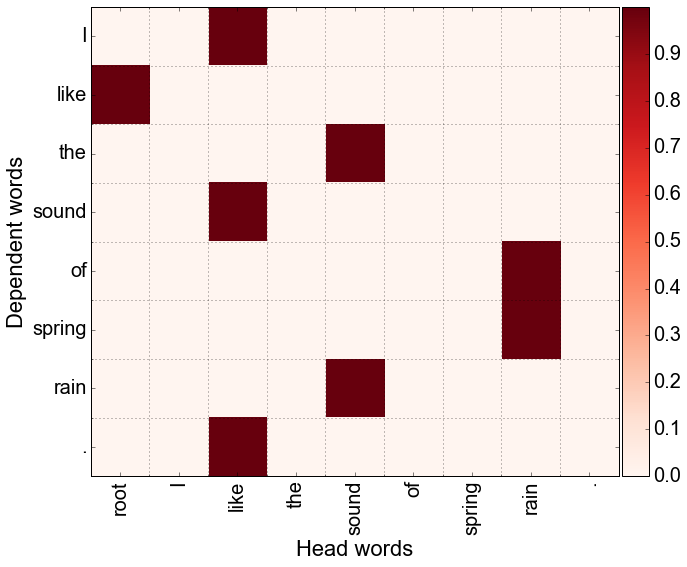}
  \caption{Probabilities assigned by the parsing network to possible
    dependency head locations. Each row represents the probabilites of
    the location of a word in the sentence.} 
  \label{fig:attention}
\end{figure}

We can make the following observation based on
Table~\ref{tab:results}: on all tested languages we can see that,
according to our expectations, we have generally that NN with
characters outerperforms NN with words. Czech and Polish belong to
morphologically reach languages, and on these languages we can observe
clear benefit from using POS-tags as a additional learning objective
(the greater role of tags in Czech and Polish is also visible, when we
look at the difference between versions with and without golden
POS tags). Furthermore, when we don't use golden tags, for Polish and
Czech our best algorithm achieves best UAS and LAS (for Polish this
remains true even when compared with MaltParser trained on gold POS tags).   

While both our neural parser and classical parsers
\cite{tiedemann_cross-lingual_2015} perform better when ground-truth
POS tags are given during inference, we observe the neural network
suffers a smaller accuracy decrease than a cascade of separately
trained tagger and parser. We hypothesize that this is due to the
end-to-end trainability of the neural parser. Comparing the scores
achieved by our parser when it has only access to word embeddings (row
``NN words'') or word spellings (row ``NN chars''), we confirm that
the reading subnetwork can extract meaningful grammatical features
from word's orthographic representations.

\paragraph{Decoding algorithm} The decoding algorithm has little
impact on the parser's performance. We have investigated the attention
outputs which show, for each word, probabilities assigned by the
network to the locations of the head word. These probability
distributions are very sharp, with virtually no ambiguities. The
attention matrix for one sentence is shown in
Figure~\ref{fig:attention}. Therefore the greedy head attachment
strategy works very well in practice.
%TODO

On the three languages tested, about 98\% of parses produced by the
greedy strategy were correct
trees, with a single ROOT and no cycles.

\section{Conclusions and Future Works}

We have presented a dependency parser that is able to operate directly
on characters, obviating the need for a traditional NLP pipeline. The
parser is trained in an end-to-end manner, and has separate cost terms
that pertain to label accuracy, head word localization and optionally
POS tagging. On morphologically rich languages the parser is competitive
with traditional transition-based solutions that use gold POS tag
information, despite the fact that no hand-designed linguistic
features are used and all information comes directly from the
orthographic form of words.

Our parser uses a distributed representation of words created by the
tagging subnetwork. In future work we plan to investigate the possibility of co-training
multilingual neural-net based parsers that permit parameter sharing
between languages to improve the models on languages with
very small corporas, such as Polish or Slovenian. A multilingual
shift-reduce parser using a set of unified POS tags was recently
proposed by \cite{ammar_many_2016} and we are curious whether the
benefits of multilingual parsing can be achieved without the manual
labor associated with the unification of linguistic features between languages.

\begin{table*}[!htbp]
  \centering
    \caption{Results of parsers trained on  UD v1.3 (newer data compared to 
    table \ref{tab:results}) 
    Our models use only orthographic representations of
    tokenized words during inference and work without a separate POS tagger.
    Ammar et al. \cite{ammar_many_2016} uses version 1.2 of UD and uses gold
    language ids and predicted coarse tags.
    SyntaxNet\cite{andor_globally_2016} works
    on predicted POS tags, while ParseySaurus\cite{alberti_parsey_saurus_2017}
    uses word spellings.}
  \label{tab:universal}
  \begin{tabular}{l l | l l | l l | c | l l}
    language & \#sentences & \multicolumn{2}{c|}{Ours} &
      \multicolumn{2}{c|}{SyntaxNet} & Ammar et al. & \multicolumn{2}{c}{ParseySaurus} \\ \hline
    & & UAS & LAS & UAS & LAS & LAS & UAS & LAS\\ \hline
    Czech & 87 913 & \textbf{91.41} & \textbf{88.18} & 89.47 & 85.93 & - & 89.09 & 84.99 \\
    Polish & 8 227 & 90.26 & 85.32 & 88.30 & 82.71 & - & \textbf{91.86} & \textbf{87.49}\\
    Russian & 5 030 & 83.29 & 79.22 & 81.75 & 77.71 & - & \textbf{84.27} & \textbf{80.65} \\
    German & 15 892 & 82.67 & 76.51 & 79.73 & 74.07 & 71.2 & \textbf{84.12} & \textbf{79.05}\\
    English & 16 622 & 87.44 & 83.94 & 84.79 & 80.38 & 79.9 & \textbf{87.86} & \textbf{84.45}\\ % Zblizamy sie do saurusa a dopiero pratraining...
    French & 16 448 & \textbf{87.25} & \textbf{83.50} & 84.68 & 81.05 & 78.5 & 86.61 & 83.1\\
    Ancient Greek & 25 251 & \textbf{78.96} & \textbf{72.36} & 68.98 & 62.07 & - & 73.85 & 68.1
  \end{tabular}
\end{table*}

\bibliography{neuraldepzotelo,neuraldep}
\bibliographystyle{icml2017}

\end{document}

%% file: drawing.pdf_tex
%% Creator: Inkscape inkscape 0.91, www.inkscape.org
%% Accompanies image file 'drawing.pdf' (pdf, eps, ps)
\begingroup%
  \makeatletter%
  \providecommand\color[2][]{%
    \errmessage{(Inkscape) Color is used for the text in Inkscape, but the package 'color.sty' is not loaded}%
    \renewcommand\color[2][]{}%
  }%
  \providecommand\transparent[1]{%
    \errmessage{(Inkscape) Transparency is used (non-zero) for the text in Inkscape, but the package 'transparent.sty' is not loaded}%
    \renewcommand\transparent[1]{}%
  }%
  \providecommand\rotatebox[2]{#2}%
  \ifx\svgwidth\undefined%
    \setlength{\unitlength}{354.33070866bp}%
    \ifx\svgscale\undefined%
      \relax%
    \else%
      \setlength{\unitlength}{\unitlength * \real{\svgscale}}%
    \fi%
  \else%
    \setlength{\unitlength}{\svgwidth}%
  \fi%
  \global\let\svgwidth\undefined%
  \global\let\svgscale\undefined%
  \makeatother%
  \begin{picture}(1,2.08)%
    \put(0,0){\includegraphics[width=\unitlength,page=1]{drawing.pdf}}%
    \put(0.22464896,0.2406141){\color[rgb]{0,0,0}\makebox(0,0)[lt]{\begin{minipage}{0.56219251\unitlength}\raggedright \end{minipage}}}%
    \put(0,0){\includegraphics[width=\unitlength,page=2]{drawing.pdf}}%
    \put(0.44241355,0.52339523){\color[rgb]{0,0,0}\makebox(0,0)[lb]{\smash{\scalebox{2}{$C^w$}}}}%
    \put(0,0){\includegraphics[width=\unitlength,page=3]{drawing.pdf}}%
    \put(0.36339136,0.67063158){\color[rgb]{0,0,0}\makebox(0,0)[lb]{\smash{$F_1$}}}%
    \put(0.45911913,0.67452509){\color[rgb]{0,0,0}\makebox(0,0)[lb]{\smash{$F_2$}}}%
    \put(0.56159124,0.6660784){\color[rgb]{0,0,0}\makebox(0,0)[lb]{\smash{$F_3$}}}%
    \put(0.55091999,0.62580152){\color[rgb]{0,0,0}\makebox(0,0)[lb]{\smash{\scalebox{0.75}{$\circledast$}}}}%
    \put(0.356767,0.62580152){\color[rgb]{0,0,0}\makebox(0,0)[lb]{\smash{\scalebox{0.75}{$\circledast$}}}}%
    \put(0.45380558,0.62608663){\color[rgb]{0,0,0}\makebox(0,0)[lb]{\smash{\scalebox{0.75}{$\circledast$}}}}%
    \put(0,0){\includegraphics[width=\unitlength,page=4]{drawing.pdf}}%
    \put(0.46379519,0.74419649){\color[rgb]{0,0,0}\makebox(0,0)[lb]{\smash{$R^w$}}}%
    \put(0,0){\includegraphics[width=\unitlength,page=5]{drawing.pdf}}%
    \put(0.46055649,0.92616656){\color[rgb]{0,0,0}\makebox(0,0)[lb]{\smash{$E^w$}}}%
    \put(0,0){\includegraphics[width=\unitlength,page=6]{drawing.pdf}}%
    \put(0.17406157,1.09111951){\color[rgb]{0,0,0}\makebox(0,0)[lb]{\smash{\scalebox{0.75}{$E^{w_0}$}}}}%
    \put(0,0){\includegraphics[width=\unitlength,page=7]{drawing.pdf}}%
    \put(0.32447559,1.09111951){\color[rgb]{0,0,0}\makebox(0,0)[lb]{\smash{\scalebox{0.75}{$E^{w_1}$}}}}%
    \put(0.47488962,1.09111951){\color[rgb]{0,0,0}\makebox(0,0)[lb]{\smash{\scalebox{0.75}{$E^{w_2}$}}}}%
    \put(0.6253036,1.09111951){\color[rgb]{0,0,0}\makebox(0,0)[lb]{\smash{\scalebox{0.75}{$E^{w_3}$}}}}%
    \put(0.77571769,1.09111951){\color[rgb]{0,0,0}\makebox(0,0)[lb]{\smash{\scalebox{0.75}{$E^{w_4}$}}}}%
    \put(0.72552484,1.40468857){\color[rgb]{0,0,0}\makebox(0,0)[lb]{\smash{\scalebox{0.75}{$H_3$}}}}%
    \put(0.22552857,1.40468857){\color[rgb]{0,0,0}\makebox(0,0)[lb]{\smash{\scalebox{0.75}{$H_1$}}}}%
    \put(0.47547824,1.40468857){\color[rgb]{0,0,0}\makebox(0,0)[lb]{\smash{\scalebox{0.75}{$H_2$}}}}%
  \end{picture}%
\endgroup%